\documentclass{article} 
\usepackage{iclr2027_conference,times}

\usepackage{amsmath,amssymb}
\usepackage{graphicx}
\usepackage{booktabs}
\usepackage{xcolor}

\usepackage{amsmath,amsfonts,bm}

\def\eqref#1{equation~\ref{#1}}

\def\1{\bm{1}}

\DeclareMathAlphabet{\mathsfit}{\encodingdefault}{\sfdefault}{m}{sl}
\SetMathAlphabet{\mathsfit}{bold}{\encodingdefault}{\sfdefault}{bx}{n}

\renewcommand{\eqref}[1]{(\ref{#1})}

\usepackage{hyperref}
\usepackage{url}

\graphicspath{{figures/}}

\newcommand{\rms}{\mathrm{rms}}

\title{The Life Cycle of a Massive Activation: Stochastic Birth, Weight-Decay-Driven\\ Growth, and Competitive Consolidation}

\author{
S. Aaron McClendon$^{1,+}$ \qquad
Jorge Gallego-Feliciano$^{2,+}$ \qquad
Antonios Saravanos$^{3,*}$\\
\\
$^{1}$Aimpoint Digital Labs, Atlanta, GA, USA\\
$^{3}$Mistral AI, Paris, FR\\
$^{3}$New York University, New York, NY, USA\\
\\
$^{*}$Corresponding author: \texttt{saravanos@nyu.edu}
}

\iclrfinalcopy

\begin{document}

\maketitle
\fancyhead{}
\begin{center}
\textbf{Code:} \url{https://github.com/apd-smcclendon/massive_activations_2}
\end{center}

\maketitle

\begin{abstract}
Massive activations, residual-stream coordinates with magnitudes far larger than typical activations, are associated with attention sinks in transformers, but how their scale is regulated during training remains incompletely understood. Combining training-trajectory analyses and controlled interventions, we trace their emergence, growth, and consolidation. Sink-carrying channels vary across random seeds but stabilize early within each run. Over longer training, surrounding channels erode and the sink concentrates onto a few redundant carriers. Across ablations, gradient attenuation follows the sink token’s collective root-mean-square magnitude rather than any single channel, making collective scale central to understanding their effects. Our central result is that weight decay causally controls the turnover of global activation scale. In controlled continuations, removing decay near the peak allows this scale to keep rising, whereas retaining it produces decline even at constant learning rate. We develop a balance model for the rise and peak of massive-activation magnitude, in which AdamW-preconditioned growth opposes weight decay. Sweeping the decay coefficient $\lambda$ shifts peak timing approximately log-linearly and yields peak magnitudes scaling approximately as $\lambda^{-1/2}$, consistent with this balance. Optimizer measurements further show that preconditioning sustains the large-channel cohort against decay even when raw maintaining forces are too small to do so. Together, these findings connect the observed life cycle to scale-regulating training dynamics and establish weight decay as a training-time lever on activation magnitude.
\end{abstract}

\section{Introduction}

A trained transformer's hidden states are not uniform. In every model examined so
far, a handful of coordinates carry magnitudes hundreds or thousands of times the
typical activation, concentrated at a single token position and largely unchanging
across inputs \cite{sun2024massive}. These massive activations are not incidental: they act as implicit attention
biases \citep{xiao2024streaming}, model quality can degrade when they are
removed \citep{yu2025superweightlargelanguage}, and they are the principal
obstacle to low-precision training and inference
\citep{dettmers2022llmint8,bondarenko2023quantizable,he2024understanding}.
That obstacle is not a fixed cost of scale: Ahmadian et al.~\cite{ahmadian2023intriguing} report that outlier dimensions are sensitive to the optimization conditions under which a model is pre-trained rather than an inherent product of its size. Park et al.~\cite{park2025osp} take that finding as a design target, replacing
Adam with an optimizer chosen to avoid privileged bases and training a 1.4B model on 1T tokens without the usual activation outliers.

Prior work gives a partial account of behaviors of these large outlier activations. \citet{chen2026gradient} supply a mechanism for
why a large magnitude is useful --- it attenuates the gradient that accumulates at
an attention sink --- in the form of a lower bound the magnitude must clear.
\citet{gallego2025hidden} show that the emergence of massive activations over
training is highly regular, fitting a five-parameter law to their growth. These are
complementary but incomplete: one is a constraint and the other a curve. Neither
says what sets the magnitude in practice, what stops it growing, or why one
coordinate rather than another comes to carry it. We address those questions, and
the answer to the last turns out to govern how the first two should be read.

\section{Related work}\label{sec:related}

\paragraph{Massive activations and attention sinks.} \citet{sun2024massive}
characterised massive activations as a small number of near-constant,
extreme-magnitude coordinates appearing at fixed feature dimensions and
predominantly at the first token or at delimiters. They function as implicit
attention biases: supplying the architecture with explicit learnable bias
parameters removes the need for them, and they do not form. The forward-side
counterpart is the attention sink \citep{xiao2024streaming} --- disproportionate
attention mass routed to an uninformative early token --- which
\citet{xiao2024streaming} exploit for streaming inference by retaining sink tokens in the KV cache. Why a model should want a sink has
since received a second answer: Barbero et al.~\cite{barbero2025first} argue that attending to the first token limits over-mixing of representations across depth, so the sink guards against representational collapse in long contexts.

\citet{darcet2023vision} report the same phenomenon in vision
A parallel, static line of work documents outlier dimensions as an obstacle to quantization~\cite{dettmers2022llmint8,kovaleva2021bert,bondarenko2023quantizable}. A further line asks why the residual stream should have a privileged basis at all, given that the architecture is close to rotation-symmetric: Elhage et al.~\cite{elhage2023privileged} rule out LayerNorm and finite-precision
arithmetic as the cause and attribute the alignment to the per-dimension
normalizers in Adam. That places the optimizer rather than the architecture at the origin of coordinate-aligned outliers, a reading our \S9 and \S11 support from the training-dynamics side.

\paragraph{The gradient-sink account.} \citet{chen2026gradient} give the mechanistic
account this paper builds on. Under causal masking, a token attended to by many
later positions also aggregates their gradients --- a gradient sink. Because the
normalisation Jacobian attenuates gradients in proportion to $1/\rms(x)$, a large
residual magnitude at that position suppresses the accumulated gradient. Their
Theorem gives a sufficient condition,
\begin{equation}\label{eq:bound}
\rms(x) \;\ge\; \frac{\lVert\gamma\rVert_\infty}{\tau}\,\lVert g_y\rVert
\;\Longrightarrow\;
\lVert\nabla_x\mathcal{L}\rVert \le \tau ,
\end{equation}
for any target gradient scale $\tau$, where $\gamma$ is the normalisation gain
vector, $g_y$ the gradient arriving at the normalisation output, and $\mathcal{L}$
the training loss. They test the implication with V-scale, a
modification attenuating gradient flow on the value path, and report that massive
activations are suppressed while attention sinks persist.

Two features of Equation~\eqref{eq:bound} shape what follows. It is stated on
$\rms(x)$, a property of the whole token rather than of any coordinate. And it is
one-sided: it establishes a floor and penalises nothing above it, so it predicts
neither a peak nor a decline during the training cycle and training dynamics.

\paragraph{Emergence over training.} \citet{gallego2025hidden} characterised the
developmental side, fitting the max-to-median ratio across 188 layers of the Pythia
suite with
\begin{equation}\label{eq:lognorm}
f(t) = A\,e^{-\lambda_{\mathrm{fit}} x_t}\log(x_t) + K,
\qquad x_t = \gamma t + t_0 ,
\end{equation}
at a mean $R^2$ of $0.984$, and predicting its parameters from architectural
features. The law has a critical point at which the ratio turns over, but the mechanism producing it was left open. On the forward side, Gu et
l.~\cite{gu2025sink} study when the attention sink appears over pre-training and how optimization, data distribution, loss and architecture bear on whether it appears at all, reading the sink as a key bias that absorbs surplus attention without contributing to the value computation. Their subject is the sink's appearance; ours is the trajectory of the massive activation that carries it, including what happens after the sink is established

\section{Background}\label{sec:method}

We study the life cycle of massive activations by combining measurements across training with controlled perturbations. Our approach pairs densely checkpointed training runs with analyses of released Pythia checkpoints. The experiments used to measure these quantities are setup according to details in Appendix \ref{app:setup}. Theory needed to understand these experiments is covered below. 

\subsection{Architecture and hidden states}

We consider decoder-only transformers of $L$ pre-norm residual blocks. Block $\ell$
receives a hidden state $h_{\ell-1}\in\mathbb{R}^{S\times d}$, where $S$ is the
sequence length and $d$ the model width, and returns
\begin{equation}
h_\ell = h_{\ell-1} + \mathcal{F}_\ell\!\left(\mathrm{Norm}(h_{\ell-1})\right),
\end{equation}
where $\mathcal{F}_\ell$ comprises multi-head self-attention and an MLP, and $\mathrm{Norm}$ is LayerNorm~\citep{ba2016layernorm} or, in the variant of \S\ref{sec:method}, RMSNorm~\citep{zhang2019rmsnorm}.

We write
$h_\ell$ for the post-residual state --- the residual stream --- and index its
entries as $h_\ell(s,c)$ for token position $s$ and channel $c$. An
\emph{activation} is one such scalar. We do not examine intermediate quantities
inside $\mathcal{F}_\ell$ except where stated.

Two consequences of the pre-norm placement matter throughout. Intermediate layers see
normalised inputs, so the forward computation depends on the \emph{direction} of
$h_\ell$ at a position far more than on its scale: a coordinate can grow very large
without much altering what the next block computes. The backward pass is different.
The Jacobian of the normalisation carries a factor of order $1/\rms$, and that
factor is not normalised away. Residual magnitude is therefore close to free in the
forward direction and consequential in the backward one --- the asymmetry that makes
large activations useful rather than merely tolerable.

\subsection{The sink token}

Attention weights at each head are a softmax over positions and therefore sum to
one. When no position is worth attending to, the mass must still be assigned
somewhere, and it accumulates at a token carrying little content --- typically the
first, which under causal masking is the only position visible to every later one,
though delimiters also serve. We write $s^\ast$ for the sink position and measure
all quantities there unless otherwise noted. We identify $s^\ast$ per model as the
position carrying the largest activations rather than assuming position $0$, since
as we show it is not always position $0$ (\S\ref{sec:stochastic}).

\subsection{Three magnitudes}

Let $x = h_\ell(s^\ast,\cdot)\in\mathbb{R}^d$ be the sink token's hidden state at
layer $\ell$ and training step $t$. We use three quantities.

\paragraph{Channel magnitude.} $m_{\ell,t}(c) = |x_c|$, the value of a single
coordinate. This is what the perturbation experiments manipulate and what the
competition among channels is fought over.

\paragraph{Ratio.} Following \citet{gallego2025hidden},
\begin{equation}\label{eq:ratio}
r_{\ell,t} = \frac{\max_c|x_c|}{\mathrm{median}_c|x_c|},
\end{equation}
the max-to-median ratio, to which Equation~\eqref{eq:emergence} was fitted. We retain
it for comparability with that work.

\paragraph{Root-mean-square.} $\rms_{\ell,t} = \sqrt{\tfrac{1}{d}\lVert x\rVert^2}$,
the token's aggregate scale, and the quantity appearing in
Equation~\eqref{eq:bound}. It is a property of the \emph{token}, not of any channel.

These three do not move together, and conflating them is the easiest way to misread
the phenomenon. In particular $r$ has a denominator that is itself changing: over
training the median falls, so $r$ can rise while the numerator $m$ of the dominant
channel is falling. We observe exactly this (\S\ref{sec:global}).

\subsection{What counts as a massive activation}

\citet{sun2024massive} define an activation as massive when $|a| > 100$ and
$|a|/\mathrm{median}(|h_\ell|)\ge 1000$. As noted in \citet{gallego2025hidden},
these absolute thresholds do not transfer to smaller models, which exhibit the same
qualitative pattern at lower values, and we follow that work in identifying the
massive activation as the largest-magnitude channel at the sink of the layer where
$r$ is greatest.


Under a threshold definition the question is settled by fiat; using the argmax instead lets us measure whether the winner is a distinct category or merely the top of a continuum. We therefore also report the \emph{isolation} of the leading channels: the largest ratio between consecutive entries of the sorted sink-magnitude profile. Isolation near~$1$ indicates a populated continuum; a large isolation indicates a small set standing apart from a collapsed bulk. We use \emph{cohort} for the leading channels above that gap.

\subsection{Gradient quantities}

The mechanism under test concerns the backward pass, so we measure gradients as well
as activations. We write $g_x$ for the gradient with respect to the sink token's representation $x$ at the input to a block's normalisation.
\citet{chen2026gradient} show that, under RMSNorm, this gradient is attenuated by
the token's own scale,
\begin{equation}\label{eq:bound}
\lVert g_x\rVert \;\le\; \frac{\lVert\gamma\rVert_\infty\,\lVert g_y\rVert}{\rms(x)},
\end{equation}
where $\gamma$ is the normalisation gain and $g_y$ the gradient at the normalisation
output. LayerNorm's Jacobian carries the same $1/\rms(x)$ factor up to centring, so
the prediction applies to both variants. Equation~\eqref{eq:bound} is the relation
our ablations test in \S\ref{sec:floor}.

A channel is written into the residual stream by particular rows of the
output-projection matrices of the blocks below it. We call these the channel's
\emph{writer rows} and collect them as $w$. A channel's magnitude grows or shrinks
according to whether the loss-driven force on $w$ exceeds the pull of weight decay,
which we quantify by the \emph{maintaining force}
\begin{equation}\label{eq:maintain}
G \;=\; \frac{-\langle w,g_w\rangle}{\lVert w\rVert^2},
\end{equation}
where $g_w$ is the gradient on the writer rows and $\langle\cdot,\cdot\rangle$
denotes the Frobenius inner product. The numerator measures how strongly the loss
pushes $w$ along its own direction; normalising by $\lVert w\rVert^2$ makes
$G$ a fractional growth rate, directly comparable to the weight-decay
coefficient $\lambda$, which enters the AdamW \citep{loshchilov2019decoupled} update as $w \leftarrow w -
\eta(g_w + \lambda w)$. The channel therefore  grows when $G > \lambda$
and erodes otherwise.

Two versions of Equation~\eqref{eq:maintain} appear below and they differ
substantially. The \emph{raw} maintaining force $G$ uses $g_w$ directly. The
\emph{preconditioned} force $\tilde G$ applies AdamW's second-moment normalisation $\tilde g_w$ \citep{kingma2015adam}, dividing
each element by the square root of its accumulated squared gradient. Because a small
but persistent gradient has a correspondingly small second moment, preconditioning
can amplify it by orders of magnitude. As we show (\S\ref{sec:birthdeath}), the raw
force on every large channel falls far below $\lambda$ --- so the sink cannot be
sustained by raw gradient magnitude at all --- while the preconditioned force does
not. Statements about what maintains a channel must therefore be made after
preconditioning; instantaneous raw gradients do not read these dynamics.

\section{Results}
\label{sec:results}

We examine how massive activations develop, how their magnitude
is regulated, and how they concentrate onto a small cohort.
We first characterise activation trajectories and the role of
weight decay in global turnover. We then use ablations and
magnitude-matched controls to distinguish effects of channel
identity from effects of magnitude and collective scale.
Finally, we examine stochastic channel identity, scale, and
training maturity, distinguishing optimiser-mediated cohort
maintenance from the unresolved mechanism of survivor selection.

\subsection{Massive activations emerge and follow a global law}
\label{sec:global}

The dominant-layer max/median ratio follows the previously reported
five-parameter law \citep{gallego2025hidden}, while sizeable channels
share a rise-and-decline pattern in raw magnitude. These quantities
need not move together: the ratio can continue rising while the MA
magnitude falls because the median decreases faster. Full trajectories,
fit results, and the qualifications on the peak-location analysis are
provided in Appendices~\ref{app:emergence} and~\ref{app:global}.

\subsection{Weight decay causes the turnover}
\label{sec:decay}

\begin{figure}[t]
    \centering
    \includegraphics[width=\linewidth]{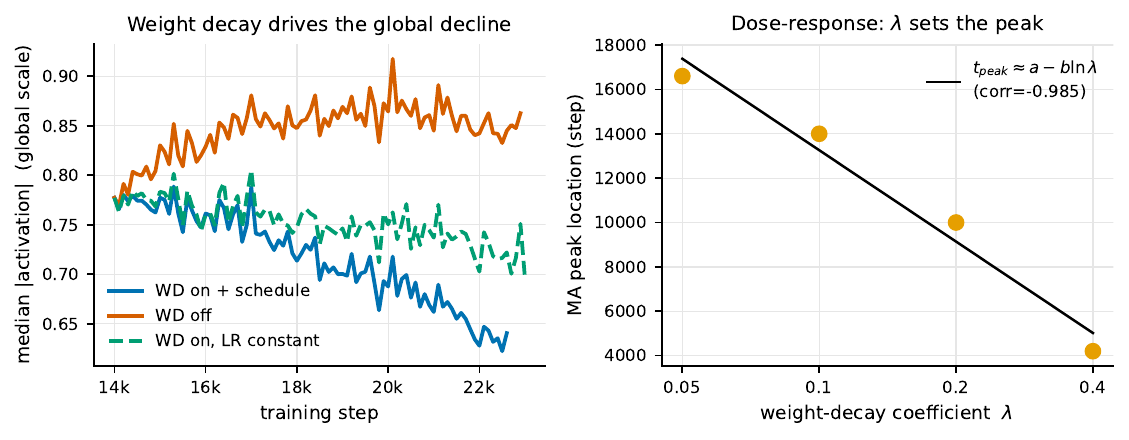}
    \caption{
        \textbf{Weight decay causes the turnover.}
        (a) Global scale (median $|x|$) rises without weight decay
        and declines with it, including at constant learning rate.
        The decaying schedule amplifies the decline.
        (b) MA peak location varies approximately log-linearly with
        the decay coefficient ($\mathrm{corr}=-0.985$).
    }
    \label{fig:decay}
\end{figure}

Continuing from a near-peak checkpoint without weight decay, the global
activation scale (median $|x|$) keeps rising. With weight decay, it
falls even at constant learning rate: $-11\%$ over the compared window,
versus $-18\%$ under the decaying schedule (Fig.~\ref{fig:decay}a).
Thus, in these continuations, weight decay drives the global turnover
and the schedule amplifies the decline. This causal comparison concerns
the global median: the individual MA is noisier and plateaus rather
than declines at constant learning rate.

Sweeping the decay coefficient under the schedule shifts the MA's peak
location approximately log-linearly,
$t_{\mathrm{peak}}\approx a-b\ln\lambda$
($\mathrm{corr}=-0.985$; Fig.~\ref{fig:decay}b).
At $\lambda=0$, no peak occurs within the observation window. Peak
magnitude scales approximately as $\lambda^{-1/2}$, consistent with
the rise-and-peak balance model in \S\ref{sec:theory}.

\subsection{The gradient floor is on the token's collective rms}
\label{sec:floor}

The gradient-sink bound predicts $\lVert g_x\rVert\propto1/\rms(x)$
(Equation~\eqref{eq:bound}). We ablate the MA and other large channels
individually, measuring $(\rms(x),\lVert g_x\rVert)$ with one forward and backward pass
per condition and no training. The product $\lVert g_x\rVert\cdot\rms(x)$ remains
approximately constant across ablation conditions: $0.94$ when the MA is
ablated, within the $0.89$--$0.95$ range spanned by ablations of the other
large channels
(Fig.~\ref{fig:controls}a). Ablating the MA changes the token's
$\rms$ by only approximately $5\%$. These measurements support
attenuation governed by the token's collective $\rms$, with the MA
one of several contributors rather than an exceptional channel.

\subsection{Apparent MA-specific regulation resolves to magnitude}
\label{sec:controls}

In perturb-and-restore experiments, an ablated MA is rebuilt
(a floor), while an over-driven MA is pulled back during decline
(a ceiling). These responses exceed those of a matched-\emph{norm}
random channel and occur independently of weight decay, initially
suggesting MA-specific regulation. However, the random channel differs
in activation magnitude, motivating magnitude-matched controls.

\begin{figure}[t]
    \centering
    \includegraphics[width=\linewidth]{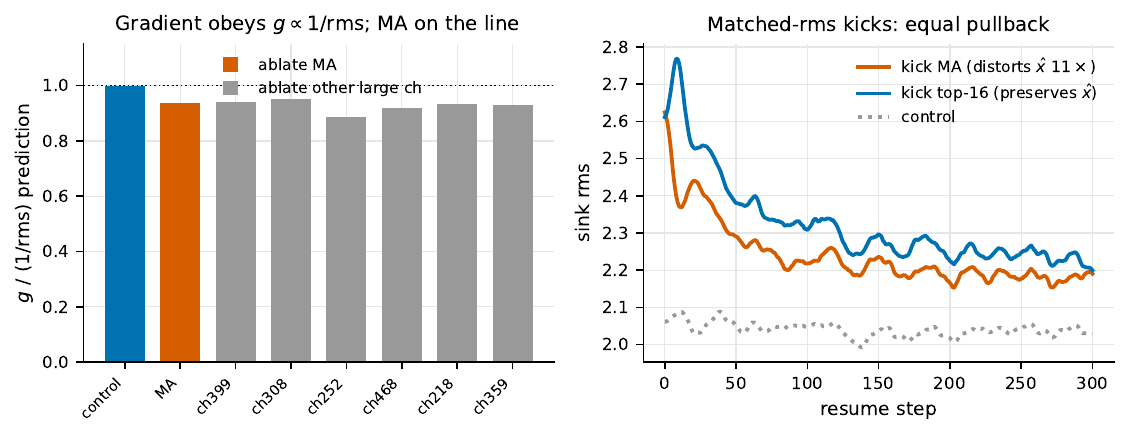}
    \caption{
        \textbf{Controls: apparent MA-specificity resolves to magnitude.}
        (a) Across individual-channel ablations, the downstream sink
        gradient follows $\lVert g_x\rVert\propto1/\rms(x)$; the MA lies with the other
        large channels.
        (b) Matched-$\rms$ displacements show similar pullback despite
        different direction distortions, supporting a collective-scale
        interpretation of the declining-regime ceiling.
    }
    \label{fig:controls}
\end{figure}

\paragraph{The ceiling tracks collective $\rms$.}
We compare scaling the MA alone with scaling the top-16 channels
proportionally, matched for the induced $\rms$ displacement.
Despite an $11\times$ difference in direction distortion, the two
perturbations remove similar fractions of the $\rms$ excess:
$73\%$ and $71\%$, respectively (Fig.~\ref{fig:controls}b).
A small random channel shows only approximately $20\%$ pullback.
The measured ceiling therefore tracks aggregate magnitude rather
than uniquely responding to the MA or its direction distortion.
These comparisons concern the continuum regime of the 51M model;
\S\ref{sec:scale} examines when comparable-magnitude peers exist.

\paragraph{Gain suppression is also a large-channel effect.}
The MA's LayerNorm gain is comparable to that of other large
channels rather than uniquely suppressed
(Appendix~\ref{app:gain}, Fig.~\ref{fig:gain}).

\subsection{The massive activation is a stochastic role, not a fixed dimension}
\label{sec:stochastic}

Changing the training seed while holding architecture and data fixed
changes the sink-carrying coordinates. In our 51M model, the dominant
channel is 207 under one seed and 263 under another; three seeds of
Pythia-410m share no top-three sink channels (Table~\ref{tab:seed}).
The number of sink channels (one to three), birth layer, and sink
position also vary across seeds. Channel identity is therefore a
stochastic outcome of training rather than a fixed architectural
coordinate.

\begin{table}[t]
    \centering
    \small
    \begin{tabular}{lll}
        \toprule
        Model & Seed & Top-three sink channels \\
        \midrule
        Ours (51M) & 1337 & 207, 399, 308 \\
        Ours (51M) & 2024 & 263, 114, 399 \\
        Pythia-410m & Default & 357, 130, 966 \\
        Pythia-410m & Seed 1 & 110, 345, 27 \\
        Pythia-410m & Seed 2 & 424, 584, 216 \\
        \bottomrule
    \end{tabular}
    \caption{
        \textbf{Sink identity varies across seeds.}
        Pythia-410m has no pairwise top-three overlap across the
        three seeds; the two 51M runs share one secondary channel.
    }
    \label{tab:seed}
\end{table}

\begin{figure}[t]
    \centering
    \includegraphics[width=\linewidth]{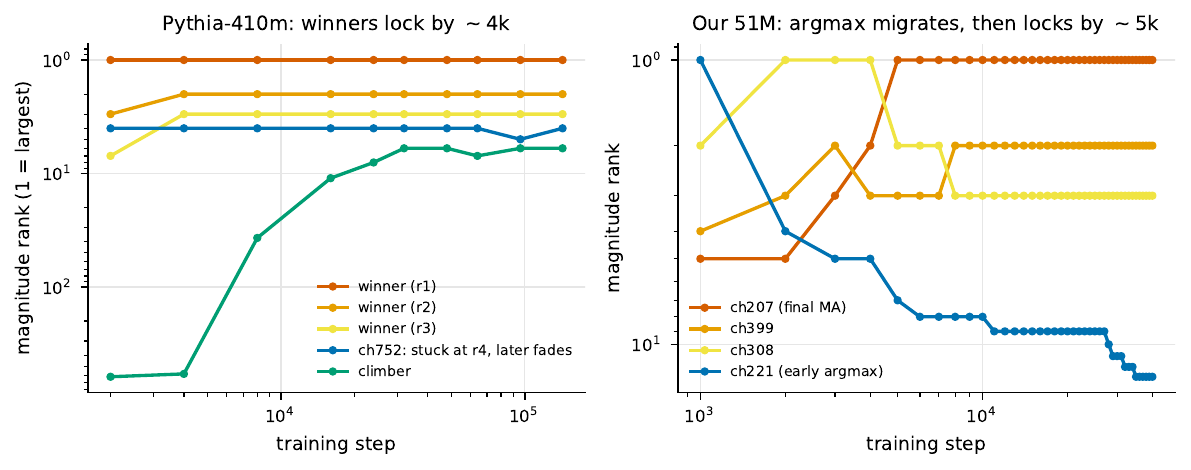}
    \caption{
        \textbf{Early lock-in of sink channels.}
        Magnitude rank versus training step (rank 1 is largest;
        log axes).
        (a) Pythia-410m's winners settle at ranks 1--3 by
        approximately 4k steps; the rank-4 channel later collapses
        (Fig.~\ref{fig:dropout}).
        (b) The 51M model's argmax migrates
        $221\to308\to207$ before locking by approximately 5k steps.
    }
    \label{fig:lockin}
\end{figure}

The identity stabilises early (Fig.~\ref{fig:lockin}). In Pythia-410m,
the eventual top three occupy ranks 1--3 from approximately step 4k
through the end of the 143k-step run. Ranks at 8k already correlate
with final ranks (Spearman $\approx0.5$--$0.7$), while a channel
remaining at rank 4 later collapses in magnitude. In our 51M model,
the argmax migrates from channel 221 to 308 to 207, then locks by
approximately step 5k. The top-three identity also remains unchanged
across ten unrelated text passages. Together, these observations
support early stochastic symmetry-breaking followed by stable,
input-robust sink identity.

\subsection{Scale and training maturity set whether the MA stands apart}
\label{sec:scale}

\begin{figure}[t]
    \centering
    \includegraphics[width=\linewidth]{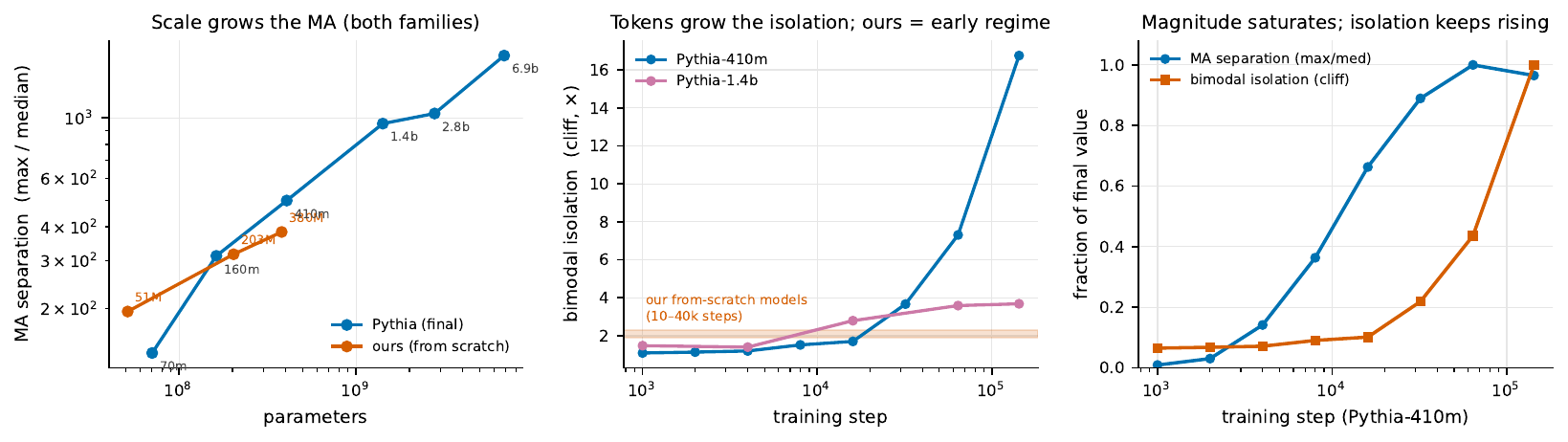}
    \caption{
        \textbf{Scale, training maturity, and isolation.}
        (a) Separation ($\max/\mathrm{median}$) versus parameter
        count. Our models fall within Pythia's range but are
        trained much less.
        (b) Isolation, the largest consecutive-rank magnitude ratio
        (\S\ref{sec:method}), across Pythia training. The shaded
        band shows our models over their 10--40k-step horizons.
        (c) Separation and isolation normalised by their final
        values in Pythia-410m: separation settles earlier, while
        isolation continues growing.
    }
    \label{fig:scale}
\end{figure}

Magnitude-matched controls require channels comparable to the MA.
We examine this condition using ranked sink-channel magnitudes
in our from-scratch models and Pythia (70M--6.9B;
Fig.~\ref{fig:scale}). Separation from the typical channel
($\max/\mathrm{median}$) increases with scale in both families,
and our models fall within Pythia's range. However, the
from-scratch models are trained much less, so this comparison mixes
scale with training maturity. Their magnitude profiles remain
populated continua, whereas fully trained Pythia models have a few
large channels separated from a collapsed bulk.

Isolation also develops during training at fixed model size.
Pythia-410m and 1.4B begin near isolation $1$, with Pythia-410m
reaching approximately $17$ by 143k steps (Fig.~\ref{fig:scale}b).
Our models remain in the low-isolation regime over their
10--40k-step training horizons, where magnitude-matched controls
are available; those comparisons do not establish the same
conclusions in the separated regime.

Separation and isolation evolve differently. In Pythia-410m,
separation largely settles by approximately $10^4$ steps, while
isolation continues increasing through the end of training
(Fig.~\ref{fig:scale}c). Magnitude growth and the emergence of a
gap between the leading cohort and the bulk are distinct processes.

\subsection{Birth and death: competitive consolidation of the sink}
\label{sec:birthdeath}

\paragraph{Concentration without conservation.}
From step 16k to 143k in Pythia-410m, the surviving channels rise
and then decline mildly, while mid-rank channels erode gradually
on staggered schedules (Fig.~\ref{fig:dropout}). The top-three's
share of the sink token's squared norm $\lVert x\rVert^2$ grows
from $0.55$ to $0.99$. This is not redistribution of a conserved
quantity: total sink $\rms$ peaks and declines, and winner gain
is approximately $2.9$ times field loss. In the 51M continuum
regime, decline is graded rather than selective: peak-to-end
retention ranges from $0.78$ to $0.89$ across the top eight
channels, with the MA part of this pattern rather than separate
from it.

\paragraph{The sink is a redundant collective.}
In Pythia-410m at step 16k, ablating a single surviving channel
reduces attention to the sink by only $0.004$--$0.015$, whereas
ablating the top three jointly produces a drop of $0.08$, an
approximately $3\times$ superadditive effect. Random-channel
ablations have negligible effect (Fig.~\ref{fig:redundancy}).
This supports a redundant implementation of the sink, consistent
with the collective-$\rms$ result (\S\ref{sec:floor}) and
attention-sink accounts \citep{sun2024massive,xiao2024streaming}.

\begin{figure}[t]
    \centering
    \includegraphics[width=0.6\linewidth]{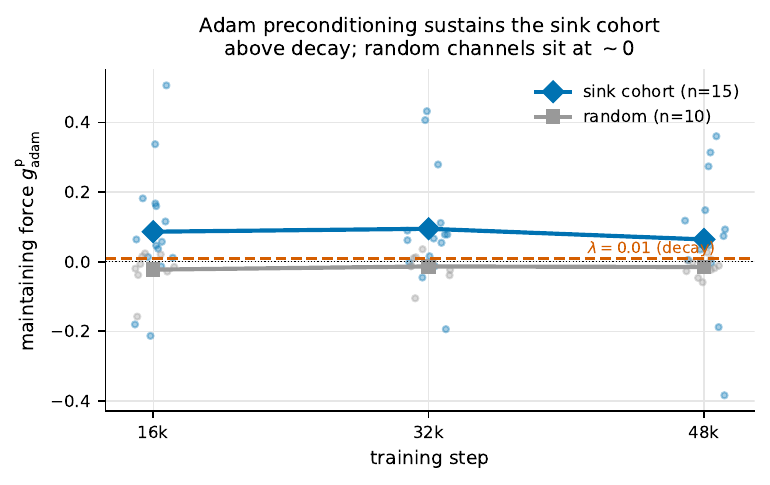}
    \caption{
        \textbf{AdamW preconditioning sustains the sink cohort.}
        The preconditioned maintaining force $\tilde G$ of
        Equation~\eqref{eq:maintain}, over three checkpoints.
        The cohort of winners and collapsers lies above Pythia's
        decay threshold $\lambda=0.01$ (dashed), while
        random channels lie near zero. Raw forces are approximately
        $10^{-4}$. Winners and collapsers overlap, leaving
        intra-cohort selection unresolved.
    }
    \label{fig:adam}
\end{figure}

\paragraph{AdamW preconditioning sustains the cohort.}
Raw maintaining forces are of order $10^{-4}$, below Pythia's
$\lambda=10^{-2}$ decay coefficient in the optimiser-force
comparison. After AdamW second-moment normalization, the
large-channel cohort lies above the decay threshold, while
random channels remain near zero (Fig.~\ref{fig:adam}).
Thus the measured cohort maintenance depends on optimiser
preconditioning rather than raw gradient magnitude. This does
not identify a mechanism selecting survivors: winners and
collapsers overlap in preconditioned maintaining force.

\paragraph{Tested mechanisms and unresolved selection.}
The measurements do not support several candidate explanations.
Collapsing channels' writer-row gradients do not fall as
$1/\rms$: the product of writer-row gradient norm and $\rms$
rises $1.5$--$2\times$, and their gradients remain comparable
to those of survivors. Across seed re-runs at 70M and 160M,
survivor count does not track total sink attention
(correlation $\approx0$), providing no support for a count set
by attention-sink saturation. Total sink $\rms$ is not conserved,
as shown above, and matched-$\rms$ perturbations do not support
an MA-specific ceiling (\S\ref{sec:controls}).

Early identity lock-in (\S\ref{sec:stochastic}) and cohort-level
maintenance do not explain why some cohort members survive and
others collapse. A survivor can even exhibit negative maintaining
force on held-out text. These post-hoc gradient measurements do
not identify an intra-cohort selector; that mechanism remains
unresolved.

\section{Discussion and Conclusion}

Together, our results describe a life cycle of stochastic birth,
decay-controlled turnover, and competitive consolidation. Sink-carrying
channels vary across seeds but stabilise early within each run; continued
training erodes the surrounding field and concentrates the sink onto a
small, redundant cohort.  Weight decay causally drives turnover in global activation scale, while the
massive activation's rise and peak are consistent with a balance between
optimiser-preconditioned growth and decay (Appendix~\ref{sec:theory}).
Its subsequent decline reflects reduced activation per unit writer-weight norm,
not necessarily shrinking weights.
AdamW's second-moment normalisation sustains the cohort against decay, but the
measured maintaining force does not distinguish eventual survivors.

The central distinction is between channel identity and collective
magnitude. Across the tested ablations, gradient attenuation follows
$\lVert g_x\rVert\propto1/\rms(x)$, locating the effect in the sink token's aggregate scale
rather than any single coordinate. In the continuum regime,
magnitude-matched controls resolve apparent MA-specific regulation into
magnitude effects. A functional role for massive activations and
magnitude-organised training dynamics are therefore compatible
descriptions at different levels.

These findings have consequences for how massive activations are interpreted
and targeted.
Because the role is stochastic — filled by whichever channels win an early
symmetry-breaking competition — interventions that zero or suppress individual
channels may leave the underlying role intact: the sink will reconstitute on
different coordinates, as the seed experiments confirm.
Because gradient attenuation is governed by the sink token's collective
\textsc{rms} rather than any privileged dimension, the effect is distributed
across the cohort and not uniquely attributable to the largest channel.
Mitigation strategies that target the token's aggregate scale, rather than
individual outliers, are therefore better motivated by the present account.

These conclusions are limited by experimental scale and training regime.
Identity and life-cycle analyses include Pythia models spanning
70M-6.9B parameters, whereas perturb-and-restore controls remain at
approximately 51M. Magnitude-matched comparisons require a populated
tail and do not establish the same behavior in the bimodal,
late-training regime. We do not test the attention-bias function
\citep{sun2024massive} or rebuilding when denied a home
\citep{vemula2026robust}; intra-cohort survivor selection also remains
unresolved. Within this scope, a massive activation is not a privileged
architectural coordinate, but a training-selected contributor to a
collective sink.

\subsection*{Ethics statement}
An institutional review process determined that this work is exempt human-subjects research; identifying details, including the institution and protocol number, are withheld during double-blind review and will be disclosed in the camera-ready version if the paper is accepted.

\subsection*{AI use statement}
The authors used generative AI tools to assist with spelling and grammar checking and with drafting and editing portions of the manuscript, including improving clarity, readability, and consistency. Generative AI tools were not used for the scientific design, execution, analysis, or interpretation of the work. All AI-assisted text was reviewed, edited, and verified by the authors. The authors take full responsibility for the final content of this work, including all claims, results, citations, and artifacts.

\bibliography{references}

@misc{yu2025superweightlargelanguage,
      title={The Super Weight in Large Language Models}, 
      author={Mengxia Yu and De Wang and Qi Shan and Colorado J Reed and Alvin Wan},
      year={2025},
      eprint={2411.07191},
      archivePrefix={arXiv},
      primaryClass={cs.CL},
      url={https://arxiv.org/abs/2411.07191}, 
}

@article{he2024understanding,
  title={Understanding and minimising outlier features in transformer training},
  author={He, Bobby and Noci, Lorenzo and Paliotta, Daniele and Schlag, Imanol and Hofmann, Thomas},
  journal={Advances in Neural Information Processing Systems},
  volume={37},
  pages={83786--83846},
  year={2024}
}

@article{darcet2023vision,
  title={Vision transformers need registers},
  author={Darcet, Timoth{\'e}e and Oquab, Maxime and Mairal, Julien and Bojanowski, Piotr},
  journal={arXiv preprint arXiv:2309.16588},
  year={2023}
}

@inproceedings{sun2024massive,
  title={Massive Activations in Large Language Models},
  author={Sun, Mingjie and Chen, Xinlei and Kolter, J. Zico and Liu, Zhuang},
  booktitle={First Conference on Language Modeling (COLM)},
  year={2024},
  note={arXiv:2402.17762}
}

@article{chen2026gradient,
  title={Attention Sinks Induce Gradient Sinks: Massive Activations as Gradient
         Regulators in Transformers},
  author={Chen, Yihong and Lin, Zhouchen and Yao, Quanming},
  journal={arXiv preprint arXiv:2603.17771},
  year={2026}
}

@article{gallego2025hidden,
  title={Hidden Dynamics of Massive Activations in Transformer Training},
  author={Gallego-Feliciano, Jorge and McClendon, S. Aaron and Morinelli, Juan
          and Zervoudakis, Stavros and Saravanos, Antonios},
  journal={arXiv preprint arXiv:2508.03616},
  year={2025}
}

@article{vemula2026robust,
  title={Massive Activations Are Architecturally Robust: A Controlled
         Scratch/Commitment Residual Stream Test},
  author={Vemula, Maruthi},
  journal={arXiv preprint arXiv:2606.20743},
  year={2026}
}

@inproceedings{xiao2024streaming,
  title={Efficient Streaming Language Models with Attention Sinks},
  author={Xiao, Guangxuan and Tian, Yuandong and Chen, Beidi and Han, Song
          and Lewis, Mike},
  booktitle={International Conference on Learning Representations (ICLR)},
  year={2024},
  note={arXiv:2309.17453}
}

@inproceedings{dettmers2022llmint8,
  title={{LLM.int8()}: 8-bit Matrix Multiplication for Transformers at Scale},
  author={Dettmers, Tim and Lewis, Mike and Belkada, Younes and
          Zettlemoyer, Luke},
  booktitle={Advances in Neural Information Processing Systems (NeurIPS)},
  volume={35},
  pages={30318--30332},
  year={2022},
  note={arXiv:2208.07339}
}

@inproceedings{kovaleva2021bert,
  title={{BERT} Busters: Outlier Dimensions that Disrupt Transformers},
  author={Kovaleva, Olga and Kulshreshtha, Saurabh and Rogers, Anna and
          Rumshisky, Anna},
  booktitle={Findings of the Association for Computational Linguistics:
             ACL-IJCNLP 2021},
  pages={3392--3405},
  year={2021},
  note={arXiv:2105.06990}
}

@inproceedings{biderman2023pythia,
  title={Pythia: A Suite for Analyzing Large Language Models Across Training
         and Scaling},
  author={Biderman, Stella and Schoelkopf, Hailey and Anthony, Quentin G. and
          Bradley, Herbie and O'Brien, Kyle and Hallahan, Eric and
          Khan, Mohammad Aflah and Purohit, Shivanshu and
          Prashanth, USVSN Sai and Raff, Edward and Skowron, Aviya and
          Sutawika, Lintang and van der Wal, Oskar},
  booktitle={Proceedings of the 40th International Conference on Machine
             Learning (ICML)},
  series={Proceedings of Machine Learning Research},
  volume={202},
  pages={2397--2430},
  year={2023},
  publisher={PMLR},
  note={arXiv:2304.01373}
}

@inproceedings{soldaini2024dolma,
  title     = {Dolma: an Open Corpus of Three Trillion Tokens for Language Model Pretraining Research},
  author    = {Soldaini, Luca and Kinney, Rodney and Bhagia, Akshita and Schwenk, Dustin and Atkinson, David and others},
  booktitle = {Proceedings of the 62nd Annual Meeting of the Association for Computational Linguistics (Volume 1: Long Papers)},
  pages     = {15725--15788},
  year      = {2024},
  doi       = {10.18653/v1/2024.acl-long.840}
}

@article{radford2019language,
  title   = {Language Models are Unsupervised Multitask Learners},
  author  = {Radford, Alec and Wu, Jeffrey and Child, Rewon and Luan, David and Amodei, Dario and Sutskever, Ilya},
  journal = {OpenAI blog},
  volume  = {1},
  number  = {8},
  pages   = {9},
  year    = {2019}
}

@inproceedings{bondarenko2023quantizable,
  title={Quantizable Transformers: Removing Outliers by Helping Attention
         Heads Do Nothing},
  author={Bondarenko, Yelysei and Nagel, Markus and Blankevoort, Tijmen},
  booktitle={Advances in Neural Information Processing Systems (NeurIPS)},
  volume={36},
  year={2023},
  note={arXiv:2306.12929}
}

@inproceedings{loshchilov2019decoupled,
  title     = {Decoupled Weight Decay Regularization},
  author    = {Loshchilov, Ilya and Hutter, Frank},
  booktitle = {International Conference on Learning Representations (ICLR)},
  year      = {2019},
  note      = {arXiv:1711.05101}
}

@inproceedings{kingma2015adam,
  title     = {Adam: A Method for Stochastic Optimization},
  author    = {Kingma, Diederik P. and Ba, Jimmy},
  booktitle = {International Conference on Learning Representations (ICLR)},
  year      = {2015},
  note      = {arXiv:1412.6980}
}

@inproceedings{kosson2024rotational,
  title     = {Rotational Equilibrium: How Weight Decay Balances Learning Across Neural Networks},
  author    = {Kosson, Atli and Messmer, Bettina and Jaggi, Martin},
  booktitle = {Proceedings of the 41st International Conference on Machine Learning (ICML)},
  series    = {Proceedings of Machine Learning Research},
  volume    = {235},
  year      = {2024},
  note      = {arXiv:2305.17212}
}

@article{elhage2023privileged,
  title   = {Privileged Bases in the Transformer Residual Stream},
  author  = {Elhage, Nelson and Lasenby, Robert and Olah, Christopher},
  journal = {Transformer Circuits Thread},
  year    = {2023},
  url     = {https://transformer-circuits.pub/2023/privileged-basis/index.html}
}

@inproceedings{gu2025sink,
  title     = {When Attention Sink Emerges in Language Models: An Empirical View},
  author    = {Gu, Xiangming and Pang, Tianyu and Du, Chao and Liu, Qian and Zhang, Fengzhuo
               and Du, Cunxiao and Wang, Ye and Lin, Min},
  booktitle = {International Conference on Learning Representations (ICLR)},
  year      = {2025},
  note      = {arXiv:2410.10781}
}

@inproceedings{barbero2025first,
  title     = {Why do LLMs attend to the first token?},
  author    = {Barbero, Federico and Arroyo, {\'A}lvaro and Gu, Xiangming and Perivolaropoulos, Christos
               and Bronstein, Michael and Veli{\v{c}}kovi{\'c}, Petar and Pascanu, Razvan},
  booktitle = {Conference on Language Modeling (COLM)},
  year      = {2025},
  note      = {arXiv:2504.02732}
}

@inproceedings{ahmadian2023intriguing,
  title     = {Intriguing Properties of Quantization at Scale},
  author    = {Ahmadian, Arash and Dash, Saurabh and Chen, Hongyu and Venkitesh, Bharat
               and Gou, Stephen and Blunsom, Phil and {\"U}st{\"u}n, Ahmet and Hooker, Sara},
  booktitle = {Advances in Neural Information Processing Systems (NeurIPS)},
  volume    = {36},
  year      = {2023},
  note      = {arXiv:2305.19268}
}

@article{park2025osp,
  title   = {Outlier-Safe Pre-Training for Robust 4-Bit Quantization of Large Language Models},
  author  = {Park, Jungwoo and Lee, Taewhoo and Yoon, Chanwoong and Hwang, Hyeon and Kang, Jaewoo},
  journal = {arXiv preprint arXiv:2506.19697},
  year    = {2025}
}

@article{ba2016layernorm,
  title   = {Layer Normalization},
  author  = {Ba, Jimmy Lei and Kiros, Jamie Ryan and Hinton, Geoffrey E.},
  journal = {arXiv preprint arXiv:1607.06450},
  year    = {2016}
}

@inproceedings{zhang2019rmsnorm,
  title     = {Root Mean Square Layer Normalization},
  author    = {Zhang, Biao and Sennrich, Rico},
  booktitle = {Advances in Neural Information Processing Systems (NeurIPS)},
  volume    = {32},
  year      = {2019},
  note      = {arXiv:1910.07467}
}
\bibliographystyle{iclr2027_conference}

\clearpage
\appendix


\section{Theory: a balance condition, and where the decline lives}\label{sec:theory}

Let $w$ be the aggregate writer weights feeding the MA channel; to leading order the
MA magnitude $m\propto\lVert w\rVert$. Under AdamW,
$w_{t+1}=w_t-\eta(\tilde g_w+\lambda w_t)$, where
$\tilde g_w \equiv \hat\mu_t/(\sqrt{\hat\nu_t}+\epsilon)$ is the \emph{preconditioned}
update --- not the raw gradient, from which it differs by one to two orders of
magnitude on these weights (\S\ref{sec:birthdeath}). Squaring and discarding the
$O(\eta^2)$ term gives
$\tfrac{d}{dt}\lVert w\rVert^2\approx-2\eta(\langle w,\tilde g_w\rangle+\lambda\lVert
w\rVert^2)$, so $\lVert w\rVert$ grows while the update's projection onto $w$ opposes
decay, and turns over when the preconditioned maintaining force of
Equation~\eqref{eq:maintain} falls to the decay coefficient:
\begin{equation}\label{eq:balance}
\tilde G(t^\ast)=\lambda .
\end{equation}
This is the growth criterion of \S\ref{sec:method} evaluated at the peak, with the preconditioned drive balancing the decay pull, an
equality complementing~\citep{chen2026gradient}. This is the single-channel form
of a general property of training under decoupled weight decay. \citet{kosson2024rotational} show that AdamW drives weight vectors towards a
rotational equilibrium in which the expected growth contributed by the update
balances the shrinkage contributed by decay, so the equilibrium scale is set by
$\lambda$ and the drive rather than by the loss alone. Their analysis concerns
units made scale-invariant by a following normalisation, whereas our $w$ are
writer rows into the residual stream and are only partially scale-invariant
under pre-norm; we therefore take \eqref{eq:balance} as calibrated by the
$\lambda$-sweep of \S\ref{sec:decay} rather than inherited from that result. What the
connection supplies is the expectation that a balance exists at all. 

\begin{figure}[t]\centering
\includegraphics[width=\linewidth]{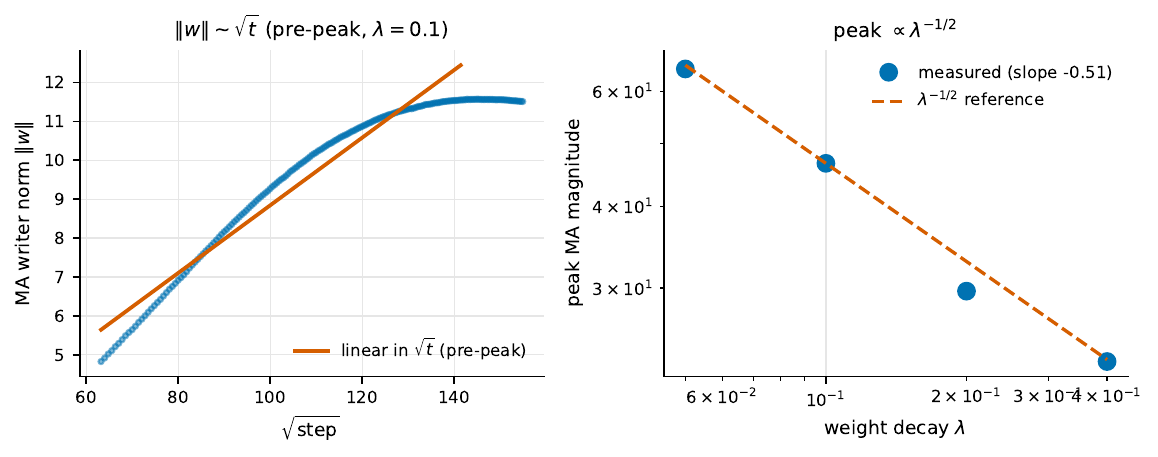}
\caption{\textbf{The balance-condition theory.} (a) The MA writer norm $\lVert w\rVert$ grows
linearly in $\sqrt{t}$ pre-peak (not exponentially), then bends over as the balance
$G=\lambda\lVert w\rVert$ is reached ($\lambda=0.1$). (b) The peak MA magnitude scales as
$\lambda^{-1/2}$ across a weight-decay sweep (measured slope $-0.51$), fixing the
drive exponent $q=-1$.}\label{fig:balance}
\end{figure}

Assuming the un-normalised preconditioned drive $\lVert w\rVert\,\tilde G = -\langle\hat w,\tilde g_w\rangle$, with
$\hat w = w/\lVert w\rVert$, follows a power law in magnitude, $\lVert w\rVert\,\tilde G = c\,m^q$, substituting into
Equation~\eqref{eq:balance} with $\lVert w\rVert\propto m$ gives $c\,m^{q-1}\propto
\lambda$, hence $m_{\mathrm{peak}}\propto\lambda^{-1/(1-q)}$ for $q\neq1$. The measured
peak-magnitude scaling ($\propto\lambda^{-1/2}$; slope $-0.51$ to $-0.57$ across
estimators, Fig.~\ref{fig:balance}b) fixes $q=-1$: the drive falls as $1/m$, so each
further increment of magnitude is worth progressively less. Because $\tilde G$ is defined on
the preconditioned update, this characterises the optimiser's effective drive rather
than the loss landscape directly; the two need not share an exponent, and we do not
assume they do. Directly measured, $\lVert w\rVert$ grows as $\sqrt t$ pre-peak (not
exponentially), the balance $G=\lambda\lVert w^\ast\rVert$ holds at the peak, and $G$
falls over training (Fig.~\ref{fig:balance}a).

\paragraph{The decline is not weight shrinkage.} At baseline weight decay the MA
magnitude turns over \emph{while its writer weights keep growing}: at $\lambda=0.1$,
$\lVert w\rVert$ rises $10.8\to11.5$ across the activation's decline while the ratio
$m/\lVert w\rVert$ falls $4.2\to3.2$. Only at strong decay does $\lVert w\rVert$
itself contract. The same weights therefore produce less activation than before, and
the decline is a change in that relationship rather than in the weights' size. We do
not claim that weight decay shrinks the MA weights. What causes the ratio to fall we
do not establish.

\paragraph{Scope of the balance law.} Equation~\eqref{eq:balance} rests on
$m\propto\lVert w\rVert$ and describes the \emph{rise and peak}, where it is
calibrated via the $\lambda$-sweep. It does \emph{not} govern the decline, precisely
because that proportionality breaks there. Two consequences follow. The
fractional-decline prediction of the single-channel $G(m)=c\,m^q$ does not carry over
to the declining, multi-channel population: the cross-channel retention ordering
during the decline (larger channels retained more; \S\ref{sec:birthdeath}) is a
competition phenomenon of that regime, not a consequence of this equation. And
$q=-1$ is fixed by the MA's \emph{own} peak scaling rather than being a universal
cross-channel law. The balance condition and the competitive consolidation are
accounts of two stages of the life cycle, and we do not conflate them.

\paragraph{Honest negatives, and their positive form.} We could not read the
regulation off instantaneous \emph{raw} gradients (activation- or weight-space,
multi-batch averaged); the signal is below chance and entangled with Adam
preconditioning. \S\ref{sec:birthdeath} turns this into a quantitative statement: the
raw maintaining gradient is one-to-two orders below the weight-decay pull at every
large channel, so the sink is held up not by raw gradient magnitude but by AdamW's
second-moment normalization---the regulation is an emergent optimiser-preconditioning
effect, visible under displacement or after preconditioning but not in the raw
gradient. A direction test did not isolate a representational cost for the ceiling.

\section{Emergence-law fit}
\label{app:emergence}

\begin{figure}[htbp]
    \centering
    \includegraphics[width=0.6\linewidth]{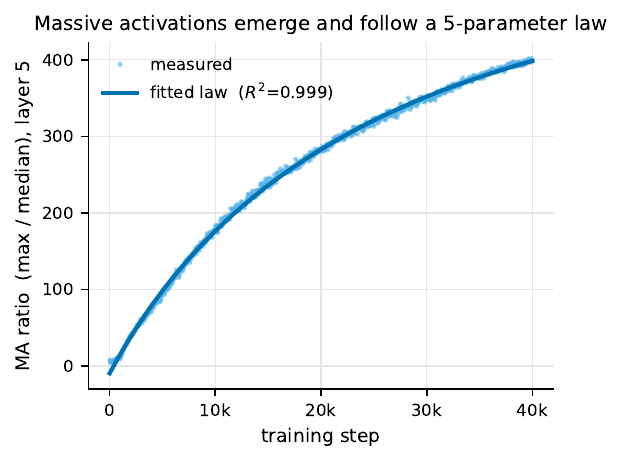}
    \caption{
        MAs emerge and fit the five-parameter law of
        \citep{gallego2025hidden}
        ($R^2=0.999$; layer 5, 51M model).
    }
    \label{fig:emergence}
\end{figure}

The max/median \emph{ratio} of the dominant layer follows the
five-parameter emergence law of \citet{gallego2025hidden},
\begin{equation}\label{eq:emergence}
r(t)=A\,e^{-\kappa x_t}\log x_t+K,\qquad x_t=\tau t+t_0,
\end{equation}
with amplitude $A$, decay rate $\kappa$, time scaling $\tau$, time offset
$t_0$ and asymptotic baseline $K$. We rename the last two from the
$\lambda,\gamma$ of \citeauthor{gallego2025hidden}, which here denote weight
decay and the normalisation gain. The fit attains $R^2=0.999$ over 40k steps
in the 51M model (Fig.~\ref{fig:emergence}).
The ratio and the raw \emph{magnitude} can disagree in direction:
at layer 5, the ratio rises monotonically to approximately $400$,
while the underlying MA magnitude peaks near step 20k and then
declines, because the median falls faster than the MA. The ratio
therefore reflects changes in both the dominant channel's magnitude
and the median, rather than the dominant channel's magnitude alone.

\section{Global channel-magnitude dynamics}
\label{app:global}

\begin{figure}[htbp]
    \centering
    \includegraphics[width=\linewidth]{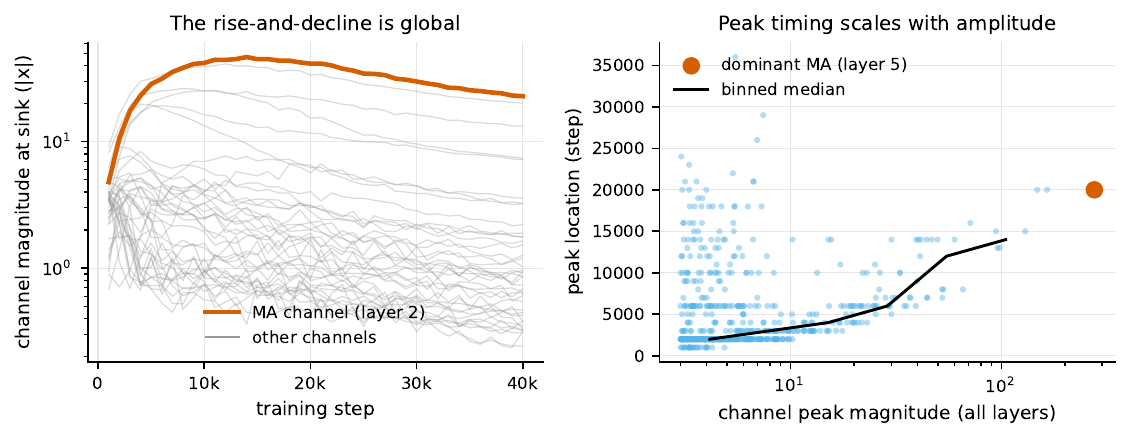}
    \caption{
        \textbf{The rise-and-decline is global.}
        (a) Per-channel sink magnitude (layer 2): nearly all
        sizeable channels rise then decline; the MA is the
        large-amplitude tail.
        (b) Peak location vs.\ channel peak magnitude, pooled over
        layers; the binned median (line) rises with amplitude, but
        the per-channel scatter is wide at any amplitude and
        \emph{right-censored}---channels peaking beyond the
        40k-step window cannot appear, biasing the high-amplitude
        end downward---so we read a tendency, not a clean law.
    }
    \label{fig:global}
\end{figure}

Measuring per-channel \emph{magnitudes} directly
(Fig.~\ref{fig:global}a, layer 2), essentially every sizeable
channel rises and then declines: 98/98 at layer 2 and 382/398 at
layer 5. The peak \emph{location} tends to increase with amplitude
(Fig.~\ref{fig:global}b), with small channels peaking early and
large channels peaking late, and the MA at the extreme.

This relationship is noisy and right-censored. Channels whose peaks
occur beyond the 40k-step observation window cannot appear in the
peak-location analysis, biasing its high-amplitude end downward.
We therefore interpret the relationship as a tendency rather than
a clean law. The rise-and-decline shape is shared across the
sizeable-channel population rather than being specific to the MA;
the five-parameter ratio law reflects both this magnitude dynamic
and the falling median.

\section{LayerNorm gain suppression}
\label{app:gain}

\begin{figure}[htbp]
    \centering
    \includegraphics[width=\linewidth]{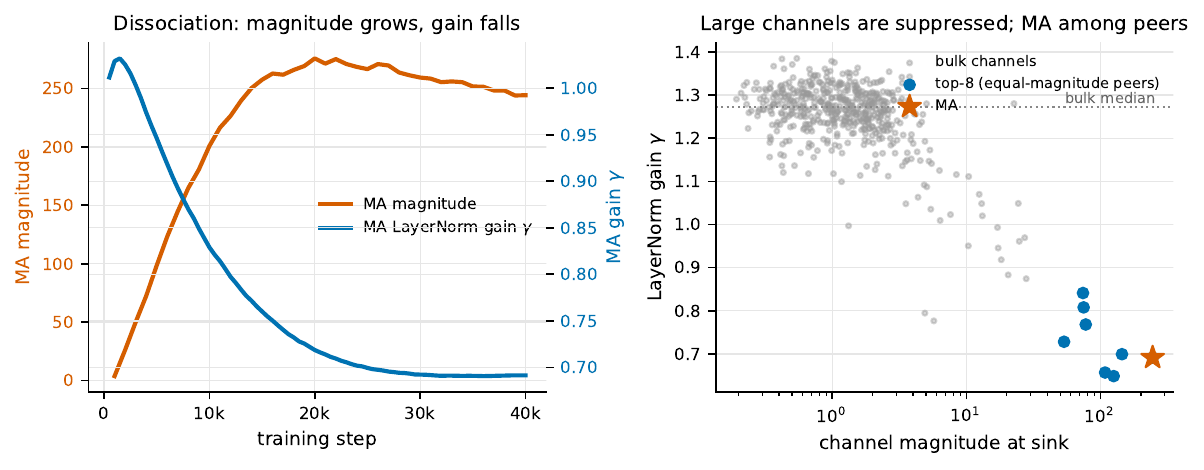}
    \caption{
        \textbf{Gain suppression tracks magnitude, not identity.}
        (a) Over training the massive activation's magnitude rises
        and holds while its LayerNorm gain $\gamma$ falls from
        approximately $1.0$ to $0.69$.
        (b) At the layer where the massive activation is largest,
        $\gamma$ decreases with channel magnitude across the
        population; the top eight channels all sit far below the
        bulk median, and the massive activation lies among them.
    }
    \label{fig:gain}
\end{figure}

On the unperturbed trajectory, the massive activation's magnitude
rises and persists while its normalisation gain $\gamma$ falls
from approximately $1.0$ early in training to $0.69$ by the end
of training (Fig.~\ref{fig:gain}a).

The suppression is not peculiar to the massive activation. At the
layer where it is largest, gain decreases with magnitude across
the whole population, and every one of the top eight channels
sits at $\gamma\approx0.65$--$0.84$, against a bulk median of
approximately $1.25$. The massive activation, at $\gamma=0.69$,
lies among its equal-magnitude peers rather than below them
(Fig.~\ref{fig:gain}b). As with the floor and the ceiling, what
looks like an MA-specific property is a property of being large.

\section{Experimental Setup}
\label{app:setup}

\subsection{Models and training}

We train decoder-only transformers from scratch at three scales, which we label by
their parameter counts: 51M (8 layers, width 512, 8 heads), 203M
(12 layers, width 1024, 16 heads), and 380M (16 layers, width 1280, 16 heads). All
use context length 512, the GPT-2 \citep{radford2019language} byte-pair vocabulary (50{,}304 tokens), no bias
terms, learned positional embeddings, and tied input/output embeddings, and are
trained on the Dolma corpus \citep{soldaini2024dolma}. The primary model uses LayerNorm, with an RMSNorm
variant at matched configuration for architecture-generality. Optimization uses
AdamW \citep{loshchilov2019decoupled} ($\beta=(0.9,0.95)$, gradient clipping at $1.0$) with weight decay
$\lambda=0.1$ and a peak learning rate of $6\times10^{-4}$ decayed by a cosine
schedule to $6\times10^{-5}$ after a linear warmup of 300 steps (400 for the two
larger models). The effective batch is 96 sequences of length 512 for the 51M model
(64 and 40 for the 203M and 380M models), in bfloat16. The three models train for
40k, 20k, and 10k steps, corresponding to approximately 2.0B, 0.7B, and 0.2B tokens.
Runs use seeds 1337 (primary) and 2024 (second seed). Checkpoints including optimiser
state are saved every 500 steps and activation probes recorded every 50 steps.

For the identity, scale, and consolidation analyses we additionally use released
Pythia checkpoints \citep{biderman2023pythia} at 70M, 160M, 410M, 1.4B, 2.8B, and
6.9B parameters, including the seed re-runs at 70M, 160M, and 410M, and the
$\sim$150 intermediate training checkpoints released per model. These are analysed
by inference only; no Pythia model is trained or fine-tuned here.

\subsection{Probes}
\label{sec:probes}
Activations are measured at the sink position of each layer over a fixed batch of 24
sequences of length 512 from the validation split, held constant across checkpoints
so that trajectories are comparable (for Pythia, a fixed passage of held-out text).
At each probe we record the per-layer ratio $r$, the $\rms$ of the sink token, the
full ranked profile of channel magnitudes --- not only the argmax, so that
near-degeneracy between the leading channels is visible --- the identity of the
argmax channel and its layer, and the norms of the writer rows for the leading
channels.

\subsection{Perturbation protocol}
\label{sec:perturb}
To ask whether a channel is actively maintained rather than passively accumulated,
we displace it and observe what training does. From a checkpoint we create forks
sharing batch order and a learning rate frozen at its value at the fork. In each
fork the target channel's writer rows are scaled by a factor $\alpha$ at every layer
--- single-layer intervention is insufficient, since the channel is written
redundantly across depth (\S\ref{sec:birthdeath}). We use $\alpha = 1$ as an
unperturbed control, which establishes how far two runs drift from data ordering
alone; $\alpha = 0$ to test for a restoring force from below; and $\alpha = 2$ to
test for one from above. Forks resume for 200--300 steps, with replicate resume
seeds where error bars are reported.

We report the perturbed fork's absolute trajectory alongside its gap to the control.
This matters because the control is itself moving during both the rising and
declining phases, so a closing gap does not by itself demonstrate that anything is
being restored. Recovery is summarised by the fraction of the initial gap closed
within the window and by the half-recovery time, which is more robust than an
exponential time constant when the return is not single-exponential.

\subsection{Controlling for magnitude}\label{sec:control-philosophy}

The perturbation protocol answers whether a channel responds; it does not answer
whether the response is peculiar to \emph{that} channel. A randomly chosen channel
differs from the massive activation in magnitude, so comparing the two confounds
identity with size. Since almost every quantity we measure varies with magnitude,
that confound is not incidental.

We therefore compare against equal magnitude wherever the comparison is available,
by three means. First, we regress the property of interest on channel magnitude and ask whether the
massive activation lies above the value its own magnitude predicts. The trend must
be fitted on channels well below the massive activation and the leading channels
scored against it out-of-sample; a trend fitted on all remaining channels is
contaminated by the runners-up, which are close in magnitude and can depress the
high end enough to manufacture an apparent outlier where none exists
(\S\ref{sec:controls}). Second, we match perturbations for the change they induce in $\rms$: a
single-channel displacement and one distributed proportionally across the leading
channels are scaled to produce the same increase in the token's aggregate magnitude,
so that the comparison isolates what distinguishes them --- how far each rotates the
token's direction --- rather than how far each moved $\rms$. Third, we pool
across layers, architectures, and seeds, and require at least 10 channels per group
before reporting any differentiator, having twice found clean separations at smaller
samples dissolve under proper sampling.

This control has a precondition, examined in \S\ref{sec:scale}: there must exist
channels of comparable magnitude to serve as the reference. Whether they do is a
property of the training regime rather than a fixed fact about transformers.

\paragraph{Weight-decay causal test (\S\ref{sec:decay}).}
Starting from a near-peak checkpoint of the 51\,M model, we resume training under two conditions sharing an identical batch order: weight decay enabled at the
checkpoint's $\lambda$, and weight decay disabled ($\lambda = 0$). The optimizer's accumulated first and second moments are carried over from the checkpoint; only the \texttt{weight\_decay} field of each parameter group is overridden. Both runs continue for 8{,}000 steps with gradient accumulation over four microbatches of 24 sequences, and activations are probed every 100 steps on the fixed validation batch.

To isolate the effect of weight decay from that of the cosine learning-rate schedule, one variant holds the learning rate fixed at the checkpoint value; the comparison in \S\ref{sec:decay} uses this constant rate fork. A second variant follows the schedule, which amplifies the decline but is not required for the qualitative result.

The dose-response curve (Fig.~\ref{fig:decay}b) is obtained by re-running the same protocol at $\lambda \in \{0.05,\,0.1,\,0.2,\,0.4\}$ under the cosine schedule and recording the training step at which the MA channel reaches its maximum magnitude in each run. Peak locations are fit to $t_{\mathrm{peak}} \approx a - b\ln\lambda$ by ordinary least squares.

\paragraph{Gradient floor test (\S\ref{sec:floor}).}
We test whether the product $\|g_x\|\cdot\mathrm{rms}(x)$ is constant across ablation conditions, as predicted by Eq.~\ref{eq:bound}, or whether ablating the MA channel produces an outlier.

\textit{Layer and channel selection.}
The target layer $L$ is the layer with the highest max/median ratio, identified automatically via the activation probe on the fixed validation batch. The MA channel is the coordinate of highest mean absolute activation at position~$s^*$ over the probe batch; six comparison channels are taken as the next-highest by the same criterion, excluding the MA.

\textit{Ablation.}
Each channel is ablated by zeroing its writer rows across all layers ($\alpha = 0$ in the notation of \S\ref{sec:perturb}), using a fresh copy of the checkpoint loaded for each condition.

\textit{Gradient extraction.}
To obtain $g_x = \partial\mathcal{L}/\partial h_L$ at the sink position, we register a forward hook on block~$L$ and call \texttt{retain\_grad()} on its
output tensor, which PyTorch would otherwise discard as a non-leaf intermediate. The model is set to training mode to enable gradient tracking, but no optimizer
step is taken. A single forward and backward pass through the cross-entropy loss is run on the
fixed probe batch of 24 sequences. We extract the residual-stream vector $x = h_L[s^*, :]$ and its gradient
$g = \partial\mathcal{L}/\partial h_L[s^*, :]$ at sink position~$s^*$, then average $\mathrm{rms}(x) = \sqrt{\tfrac{1}{d}\|x\|^2}$ and $\|g\|$ over the batch.

\textit{Test statistic.}
For each ablation condition $i$ we report $g_i \cdot \mathrm{rms}_i$ (constant under the pure $1/\mathrm{rms}$ law) and the ratio $g_i\,/\,(g_0\,\mathrm{rms}_0 / \mathrm{rms}_i)$, where subscript~0 denotes the unablated control. A ratio near 1.0 across all conditions indicates that attenuation is governed by the collective token scale rather than the identity of any individual channel.

\paragraph{Apparent MA-specific regulation (\S\ref{sec:controls}).}
Section~\ref{sec:controls} reports two classes of perturbation: a floor
experiment (ablated MA is rebuilt by continued training) and a ceiling
experiment (over-driven MA relaxes back), followed by a direction-controlled
comparison that resolves the apparent specificity to a magnitude effect.

\textit{Floor and ceiling (perturb-and-restore).}
Following the protocol of \S\ref{sec:perturb}, we create three forks from each
checkpoint: an unperturbed control ($\alpha = 1$), a perturbed MA fork
($\alpha = 0$ for the floor; $\alpha = 2$ for the ceiling), and a fork
perturbing a random channel matched by writer-row L2 norm to the MA.
All forks share batch order and resume with the learning rate frozen at
its checkpoint value.
Forks run for 200--300 steps and recovery is summarised by the fraction of the
initial perturbation gap closed within the window and the model-free half-recovery
time $t_{1/2}$.

\textit{Ceiling with matched-norm random channel.}
For the ceiling comparison, the random channel is selected as the channel whose
aggregate writer-row L2 norm $\|\mathbf{w}\|$ is closest to the MA's, excluding
the MA itself.
Both the MA and the random channel are scaled by $\alpha = 2$ in separate forks.
Recovery is measured as the fraction of the initial excess magnitude
$(m_0 - m_0/\alpha)$ removed by the end of the window:
$(m_0 - m_{\mathrm{end}})/(m_0 - m_0/\alpha)$.

\textit{Direction-controlled comparison (Fig.~\ref{fig:controls}b).}
To test whether the ceiling reflects MA identity or aggregate token scale, we
compare two perturbations from the same checkpoint that induce equal
$\mathrm{rms}$ displacement at the sink but differ in direction distortion by a
factor of $11\times$:

\begin{itemize}
\item \textbf{Kick A}: scale the MA channel alone by $\alpha = 2$, strongly distorting the sink direction.
\item \textbf{Kick B}: scale the top-16 channels by a factor $\beta$, calibrated iteratively (up to three passes, convergence tolerance 3\%) so that the resulting $\mathrm{rms}(x)$ matches Kick~A, preserving the sink direction.
\end{itemize}

Both forks resume for 300 steps.
We track $\mathrm{rms}(x_{\mathrm{sink}})$ and the cosine distortion
$1 - \cos(\hat{x}_t,\,\hat{x}^{\mathrm{ctrl}}_t)$ at each step, where
$\hat{x}^{\mathrm{ctrl}}_t$ is the unit sink vector of the unperturbed control
at the same step.

\paragraph{Scale and isolation profiles (\S\ref{sec:scale}).}
Section~\ref{sec:scale} reports two measurements: a cross-model snapshot at
the end of training (Fig.~\ref{fig:scale}a), and a trajectory of isolation
over training in Pythia-410m (Fig.~\ref{fig:scale}b--c).

\textit{Cross-model scale profile.}
For the 51\,M from-scratch model we use the final checkpoint (step~40k) with
the fixed probe batch (\S\ref{sec:probes}), hooking all layers simultaneously
to obtain residual-stream activations.
For Pythia models (70M--6.9B) we load the final released weights from
Hugging Face, pass a single fixed passage of 512 tokens (batch size~4) through
\texttt{output\_hidden\_states}, and collect all layer outputs in one forward
pass.
In both cases the target layer is identified automatically as the layer with
the highest max/median ratio across all layers and token positions.

We report two complementary measures of how far the leading channels stand
apart from the rest:
\begin{itemize}
\item \textbf{Separation} (max\,/\,median): the ratio of the largest channel magnitude to the median across channels at the sink position, following Eq.~\ref{eq:ratio}.
\item \textbf{Isolation} (cliff): the largest ratio between adjacent entries
in the top-15 sorted magnitude profile,
$\max_{i=0}^{13}\,(s_i / s_{i+1})$,
where $s_0 \geq s_1 \geq \cdots$ are the sorted magnitudes.
A large cliff indicates that the leading cohort has pulled away from the
remainder; a cliff near 1 indicates a populated continuum.
\end{itemize}

\textit{Isolation trajectory (Pythia-410m).}
We load Pythia-410m at checkpoints
$\{1\mathrm{k},\,2\mathrm{k},\,4\mathrm{k},\,8\mathrm{k},\,16\mathrm{k},\,
32\mathrm{k},\,64\mathrm{k},\,143\mathrm{k}\}$ steps,
passing the same fixed passage (batch size~4, sequence length~512) at each.
Separation and isolation are computed identically to the cross-model profile
above, with the target layer re-identified independently at each checkpoint.
Figure~\ref{fig:scale}c normalises each quantity by its value at the final
checkpoint to place them on a common scale.

\paragraph{Competitive consolidation (\S\ref{sec:birthdeath}).}
Section~\ref{sec:birthdeath} reports three separate measurements on
Pythia-410m: a retention analysis of channel magnitudes over training, an
attention-ablation test of redundancy, and an optimizer-state measurement of
cohort maintenance.
All three use the released Pythia-410m checkpoints loaded from Hugging Face.

\begin{figure}[t]
    \centering
    \includegraphics[width=\linewidth]{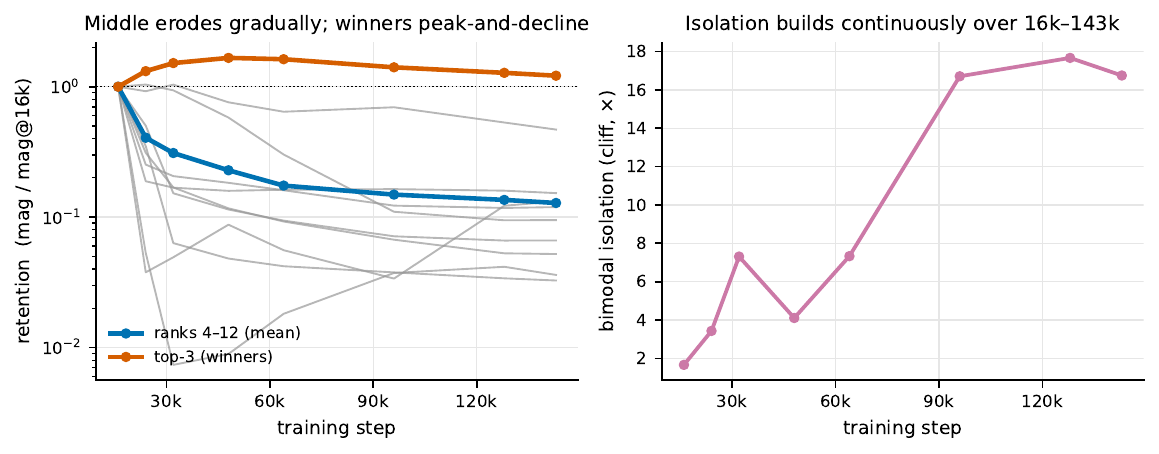}
    \caption{
        \textbf{Competitive consolidation in Pythia-410m
        (16k--143k steps).}
        (a) The top-three winners rise and then decline mildly,
        while mid-rank channels erode on staggered schedules.
        (b) Isolation builds continuously rather than appearing
        at a single transition step.
    }
    \label{fig:dropout}
\end{figure}

\textit{Retention and concentration (Fig.~\ref{fig:dropout}).}
Starting from the checkpoint at step~16k (near the emergence onset), we track
the magnitude of each channel at the sink position across checkpoints at
steps $\{16\mathrm{k},\,30\mathrm{k},\,60\mathrm{k},\,90\mathrm{k},\,
120\mathrm{k},\,143\mathrm{k}\}$.
Retention for rank~$k$ at step~$t$ is defined as
$\mathrm{mag}(t)\,/\,\mathrm{mag}(16\mathrm{k})$,
normalising each channel's trajectory to its value at the reference checkpoint.
We report the mean retention for the top-3 winners and for the mid-tier
(ranks~4--12 at step~16k) separately.

\begin{figure}[t]
    \centering
    \includegraphics[width=0.46\linewidth]{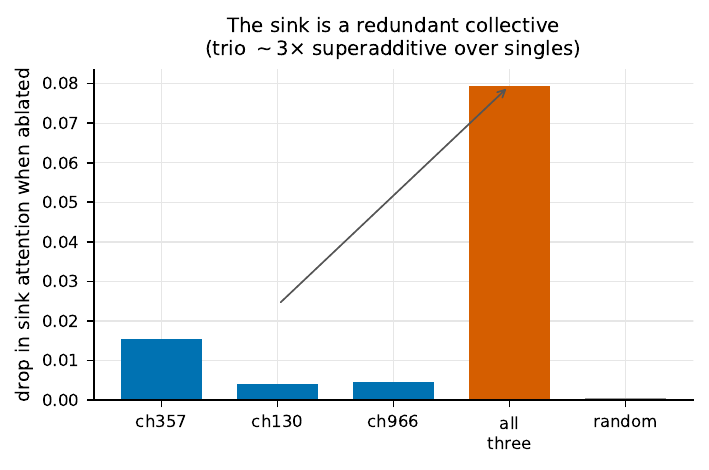}
    \caption{
        \textbf{The sink is a redundant collective.}
        Drop in sink attention after channel ablation
        (Pythia-410m, step 16k). Individual surviving-channel
        ablations have small effects; joint top-three ablation
        is approximately $3\times$ superadditive. A random-channel
        ablation has negligible effect.
    }
    \label{fig:redundancy}
\end{figure}

\textit{Redundancy test (Fig.~\ref{fig:redundancy}).}
To measure how much each channel contributes to the attention sink, we ablate
channels by zeroing them via forward hooks registered on the embedding layer and
on every transformer layer output, so that the ablated channel carries no
signal at any depth.
We then measure the mean attention weight assigned to position~0 from all
query positions $s \geq 1$, averaged across all heads and all layers, on a
single 256-token sequence.
Survivors (channels 357, 130, 966) and a collapser (channel 752) are ablated
individually and jointly; three randomly sampled channels serve as a baseline.
Superadditivity is the ratio of the joint top-three drop to the sum of the
individual drops.

\textit{AdamW preconditioning (Fig.~\ref{fig:adam}).}
We measure the preconditioned maintaining force
$\tilde{G} = -\langle \mathbf{w},\, \hat{g}_w \rangle\,/\,\|\mathbf{w}\|^2$
(Eq.~\ref{eq:maintain}) for three groups of channels at steps 16k, 32k, and 48k.
Channel groups are defined from the step-16k and step-143k checkpoints:
\textit{winners} are the top-10 channels by magnitude at step~143k;
\textit{collapsers} are channels in the top-40 at step~16k whose retention
$\mathrm{mag}(143\mathrm{k})/\mathrm{mag}(16\mathrm{k}) < 0.2$ and that
are not winners; \textit{random} channels are drawn uniformly from the
remainder.
At each checkpoint we accumulate per-sample gradients over 200 batches of 128
tokens from WikiText-2, computing the mean gradient $m$ and mean squared
gradient $v$ per writer-row entry.
The preconditioned gradient is $\hat{g}_w = m\,/\,(\sqrt{v} + \varepsilon)$,
approximating AdamW's second-moment normalisation.
Writer rows are the rows of the attention output projection
(\texttt{attention.dense.weight}) and the MLP down-projection
(\texttt{mlp.dense\_4h\_to\_h.weight}) indexed by channel.
The decay threshold $\lambda = 0.01$ is Pythia's weight-decay coefficient.

\end{document}